\pdfoutput=1
\documentclass[11pt, letterpaper, shortlabels]{berkeley}

\usepackage{hyperref}
\usepackage{color-edits}
\usepackage{wrapfig}
\usepackage{multirow}
\usepackage{colortbl}
\usepackage{xspace}
\usepackage{natbib}
\usepackage{amsmath}  
\usepackage{caption}
\usepackage{soul}
\usepackage{array}  
\usepackage{algorithm}
\usepackage{algpseudocode}

\newcommand{\etc}{{e.t.c.}}
\newcommand{\ie}{{i.e.}}

\ifx\assumption\undefined
\newtheorem{assumption}{Assumption}
\fi

\newcommand{\mas}{{BookWorld}\xspace}

\newcommand{\customcite}[2]{\hyperlink{cite.#2}{#1}}
\newcommand{\dq}[1]{``#1''}

\newcommand{\redemph}[1]{\textcolor{red}{\textbf{#1}}}
\newcommand{\greenemph}[1]{\textcolor{teal}{\textbf{#1}}}
\makeatletter                                       
\newenvironment{chapquote}[3][2.7em]
  {\setlength{\@tempdima}{#1}
   \ifx\relax#2\relax\setlength{\@tempdimb}{#1}\else\setlength{\@tempdimb}{#2}\fi
   \def\chapquote@author{#3}
   \parshape 1 \@tempdima \dimexpr\textwidth-\@tempdima-\@tempdimb\relax
   \itshape}
  {\newline\par\normalfont\hfill--\ \chapquote@author\hspace*{\@tempdimb}\par\bigskip}
\makeatother

\title{\mas: From Novels to Interactive Agent Societies for Creative Story Generation} 

\author[1]{Yiting Ran*}
\author[1]{Xintao Wang*}
\author[1]{Tian Qiu}
\author[1]{Jiaqing Liang}
\author[1]{Yanghua Xiao}
\author[1]{Deqing Yang}

\affil[*]{Equal contributions}
\affil[1]{Fudan University}

\begin{abstract}
\textbf{Abstract:} 
Recent advances in large language models (LLMs) have enabled social simulation through multi-agent systems.  
Prior efforts focus on agent societies created from scratch, assigning agents with newly defined personas.
However, simulating established fictional worlds and characters remain largely underexplored, despite its significant practical value. 
In this paper, we introduce \mas, a comprehensive system for constructing and simulating book-based multi-agent societies.
\mas's design covers comprehensive real-world intricacies, including diverse and dynamic characters, fictional worldviews, geographical constraints and changes, \etc. 
\mas enables diverse applications including story generation, interactive games and social simulation, offering novel ways to extend and explore beloved fictional works. Through extensive experiments, we demonstrate that \mas generates creative, high-quality stories while maintaining fidelity to the source books, surpassing previous methods with a win rate of 75.36\%. 
The code of this paper can be found at the project page: \href{https://bookworld2025.github.io/}{https://bookworld2025.github.io/}

\end{abstract}

\begin{document}
\maketitle

\begin{figure}[h]
    \includegraphics[width=0.95\linewidth]{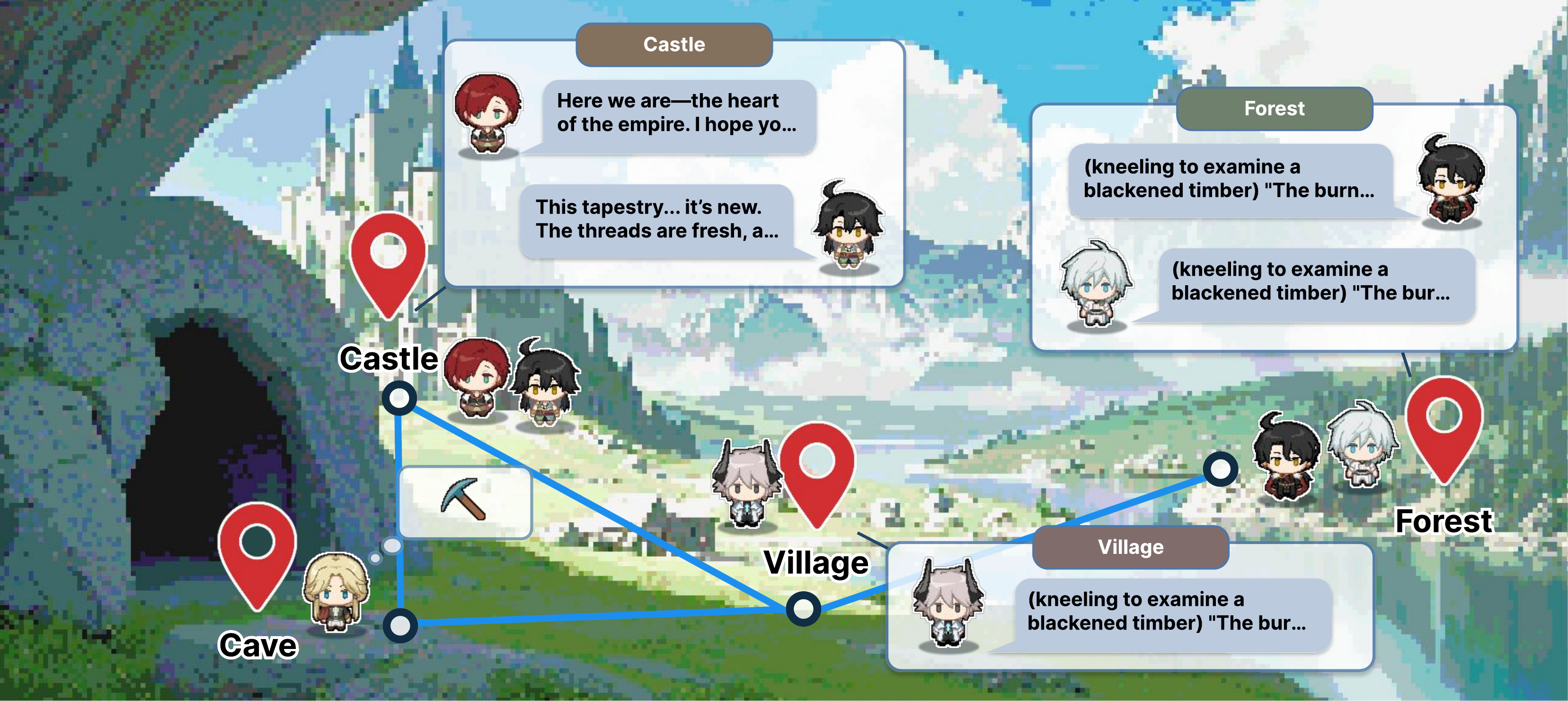}
    \label{fig:preview}
    \caption{A preview of \mas. Characters in the same location can interact with each other based on their goals and other factors. The map is represented as a discrete graph, where characters can traverse between adjacent locations spending several turns.}
\end{figure}

\vspace{-0.5cm}
\section{Introduction}

\begin{chapquote}{50pt}{\normalsize{\customcite{\textit{Sophie's World}}{gaarder1994sophie}, by Jostein Gaarder.}}
\dq{Now we know that we are living our lives in a book.}
\newline
\dq{If what you say is true, I’m going to run away from the book and go my own way.}
\vspace{-0.5cm}
\end{chapquote}

Characters in fictional works, once established, should take on a life of their own, beyond the manipulation of their authors. 
Bringing fictional characters and worlds into life has long captivated the imagination of fiction enthusiasts.
Leveraging recent advances in large language models (LLMs), we can now craft interactive artificial society for characters in books through multi-agent systems~\citep{park2023generative}, 
which simulate social interactions of humans and hence facilitate diverse applications such as interactive games~\citep{wu2024role}, social simulations~\citep{zhou2023sotopia} and story creation~\citep{han2024ibsen,chen2024hollmwood}.  

Previous efforts primarily focus on creating agent societies from scratch, where agent personas are newly defined by brief descriptions or demographic traits, including:  
\textit{1)} social simulations that study agents' social behaviors, ranging from specified scenarios such as debates~\citep{chan2023chateval} and strategic games~\citep{wang2023avalon} to open-ended society simulation~\citep{park2023generative,dai2024artificial}; and 
\textit{2)} task-oriented multi-agent systems, such as collaborative coding~\citep{huang2023agentcoder} and story generation~\citep{han2024ibsen}, where agents are specialized for different subtasks.
However, agent societies simulating established fictional worlds remain underexplored. 

In this work, we propose \mas, a comprehensive system for 
book-based multi-agent societies, which simulates story progression in fictional worlds and facilitates story creation. 
\mas extracts character data and background knowledge from source books, 
and constructs a multi-agent system using these data, comprising role agents for the characters and a world agent for simulation control.
The simulation progresses through individual scenes, where role agents of involved characters engage in various interactions such as working, communicating and trading. 
They continuously update their memories, status, and goals. 
The world agent orchestrates the simulation by managing system workflow, maintaining global status, providing environmental feedback, \etc
When the simulation ends, its histories weave together the stories, which are then polished by LLMs into cohesive, novel-style narratives. 
Our system also supports human intervention, \ie, controlling the simulation via user-specified plots or scripts.  

In addition, we systematically collect worldview data from books to enhance fictional world simulation. 
As fictional works often contain various fantastical background elements, we extract comprehensive worldview data to enrich \mas, including social norms, cultural contexts, and terminology explanations. 
These data enable agents in \mas to act appropriately under corresponding worldviews. 

The main contributions of this work are as follows:
\begin{itemize}
    \item To the best of our knowledge, this work presents the first study of book-based agent societies for fictional world simulation. 
    Such simulation allows character-driven storytelling, thus facilitating creative story generation. 
    
    \item We introduce \mas, a comprehensive framework for constructing and simulating book-based agent societies, covering systematic methodologies for data preparation, simulation, and rephrasing.
 
    \item We evaluate \mas through comprehensive quantitative and qualitative analyses. The results demonstrate that \mas generates high-quality narratives while maintaining fidelity to the source materials, outperforming previous methods in 75.36\% cases.
    
\end{itemize}
\begin{figure*}
    \centering
    \includegraphics[width=\linewidth]{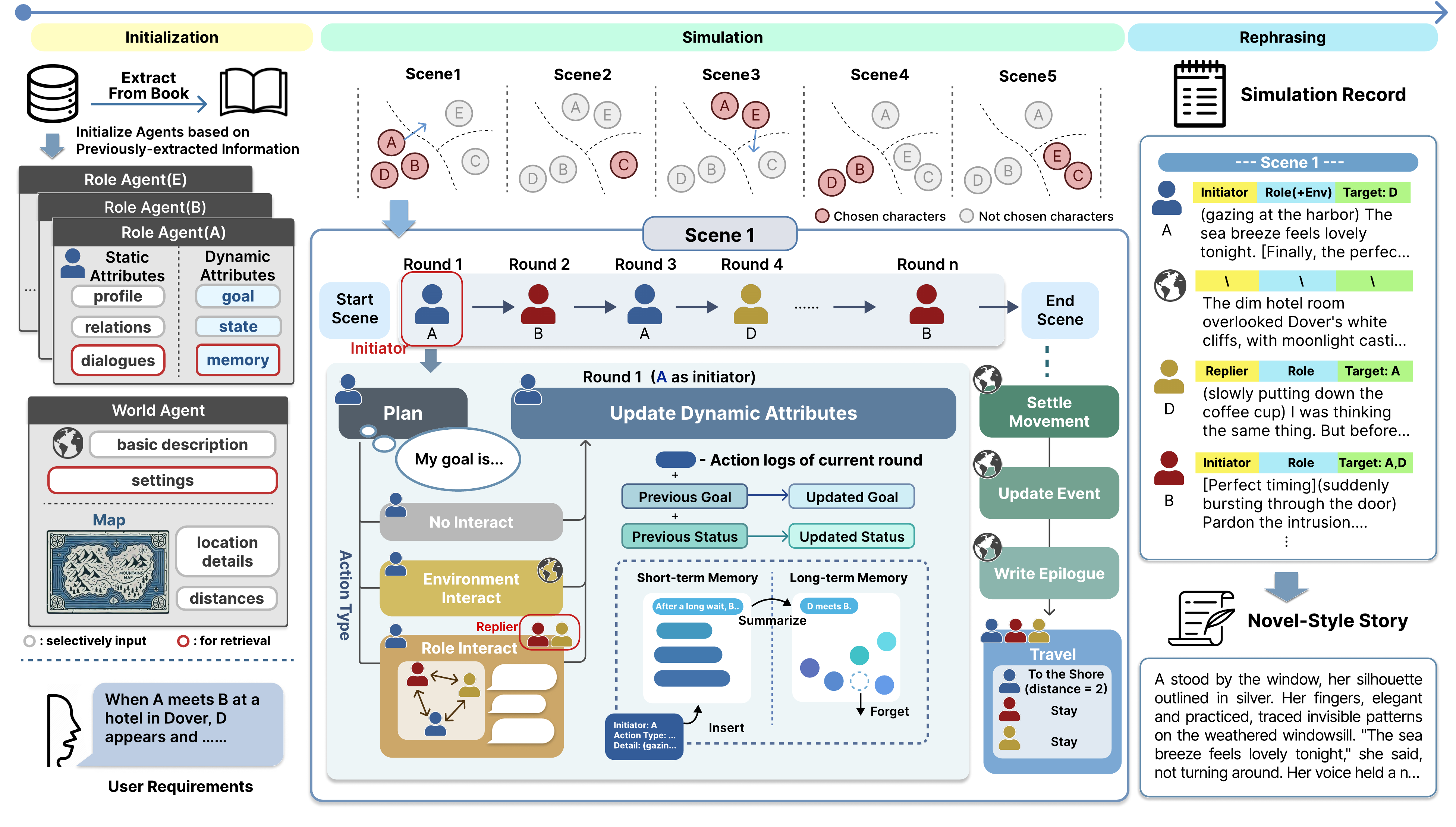}
    \caption{The complete pipeline of \mas simulation. The framework starts from initializing the agents, then the agents interact across scenes.}
    \label{fig:pipeline}
\end{figure*}

\section{Related Work}

\subsection{Multi-agent System}
\paragraph{World Simulation} Multi-agent systems for world simulation are developed to explore social dynamics in real-world applications. Such simulations could test social science theories within a small scope~\citep{chuang2024wisdom} or populate virtual spaces and communities with large-scale realistic social phenomena.

\citet{park2023generative} develops generative agents within a Sims-like interactive sandbox environment, enabling users to engage with a small community of 25 agents through natural language interactions. 
Project Sid~\citep{al2024project} deploys nearly a thousand agents with distinct personalities in Minecraft, establishing an "AI civilization" with complex human social activities such as trading and elections. \citet{yang2024oasisopenagentsocial} develops OASIS, a large-scale social media simulation framework capable of orchestrating interactions among millions of agents.

\paragraph{Story Generation}
Some works have made the efforts to leverage multi-agent systems for story generation~\citep{chen2024hollmwood, han2024ibsen}. In these works, stories are generated by writing and expanding pre-defined outlines through constructing author (or director), editor, and actor agents to collaboratively solve the story generation task.

While these works demonstrate significant potential in world simulations and story generation, there remains limited exploration of book-based virtual environment construction and its application to story creation.

\subsection{Automated Story Writing}
Early research methods emphasize algorithm planning based on character traits and social constraints~\citep{meehan1977tale, lebowitz1984creating}. With the rise of neural networks, research has gradually shifted towards data-driven machine learning methods~\citep{yao2019plan, goldfarb2020content,kreminski2021winnow}, combining or sifting stories.

The advent of large language models has significantly transformed the landscape of automated story writing~\citep{yuan2022wordcraft}. Most contemporary approaches employ a top-down methodology, generating narratives from the perspective of the writer or director, first creating high-level story outlines and then incrementally developing detailed content to enhance the overall quality of the narrative generation. While some approaches rely on a single large language model~\citep{mirowski2023co}, others leverage multi-agent architectures to generate stories~\citep{han2024ibsen, chen2024hollmwood}.

However, experimental results show that these approaches still lag behind human professional writers in intrinsic creativity and textual complexity~\citep{tian2024large}, often generating stories that lack suspense and tension and tend to produce homogenized, creatively lacking content \citep{gomez2023confederacy, ismayilzada2024evaluating}.

\section{\mas}

In this section, we elaborate on the design of \mas.
The primary motive of \mas is to build an interactive system for book-based multi-agent simulation, which includes two major objectives: 
\begin{itemize}
    \item Providing immersive experiences where character agents (and users) feel as if they are situated in the fictional world; 
    \item Enhancing character autonomy within the system while preserving fidelity to their established personality and experiences.
\end{itemize}

Towards these objectives, we concentrate on the design principles of data and agents.
The original text is transformed into structured data through a well-designed extraction method, which filters out the key information used to build the agents.
In addition, we innovatively develop a specialized method to collect worldview details, ensuring adherence to global norms in fictional worldviews. 

In \mas. the character agents are granted substantial autonomy. 
Their actions include exploring the environment, interacting with other characters, and responding to various stimuli. 
A dynamic attributes updating mechanism allows them to reflect on their experiences throughout simulations, all while preserving character fidelity and consistency.

\subsection{Overview}
The overall simulation pipeline is illustrated in Figure~\ref{fig:pipeline}. 

\paragraph{Data Preparation}
Before the simulation starts, we extract character and worldview data from source materials, used for book-to-system construction. The extraction method is detailed in \S\ref{sec:data_preparation}.

\paragraph{Simulation}
The simulation begins by initializing role agents and the world agent, loading character profiles, the geospatial  map, the worldview data and other necessary information.
Each character establishes long-term motivations representing fundamental aspirations, such as \textit{defending my country}. Details about the agents are introduced in \S\ref{sec:architecture}.

Upon completing initialization, the simulation phase commences. Referring to dramatic theory~\citep{mckee1999story}, we define the minimal narrative unit as a \textbf{scene} to preserve narrative integrity and coherence. A scene is a bounded segment, analogous to a chapter in novels. The simulation continues for a user-specified number of scenes. Scenes retain modular independence, and they collectively compose an integral, cohesive narrative.

Within each scene, a group of selected role agents act one after another. 
They execute actions, interact with other agents or the environment, driven by their personal goals. 
Detailed scene arrangements are described in \S\ref{sec:simulation_implementation}. 

\mas incorporates comprehensive geospatial modeling, where the system tracks and updates role agents' locations dynamically. 
As characters navigate through the fictional world, their movements are governed by geographical constraints and travel times. When agents decide to travel, they must spend several scenes in transit before reaching their destinations. This process is orchestrated by the world agent, which also select scene participants based on their current whereabouts and historical interactions.

\paragraph{Rephrasing} After the simulation ends, we collect the simulation records and apply LLMs to rephrase the records into the final, novel-style story.

\subsection{Architecture}
\label{sec:architecture}
Our approach empowers characters to act within a spatially and culturally grounded virtual world, enabling stories to emerge naturally from their interactions. To achieve this, we introduce two core components: \textbf{role agents} and \textbf{world agent}. These components are designed to work together, bridging individual character motivations with global environmental constraints, thereby ensuring both narrative creativity and logical coherence.

\subsubsection{Role Agents}
Role agents are the core of \mas. They are able to make actions based on their intrinsic traits and exhibit complex social behaviors, forming individual motivations and memories during the simulation. Those agents are equipped with a long-term memory module based on retrieval augmentation.

\paragraph{Attributes}
Role attributes contain fundamental traits essential for agent construction. These attributes are categorized into static and dynamic types:

\begin{itemize}
    \item{\textbf{Static Attributes}} Static attributes include inherent characteristics such as gender, age, appearance, and personality. These attributes, specified in character profiles, remain constant throughout the simulation to maintain character consistency.
    \item{\textbf{Dynamic Attributes}} Dynamic attributes mainly include goals, states and memories. Initialized at the beginning, these attributes evolve with story progression, enabling dynamic character development.
\end{itemize}

\paragraph{Actions}
\label{sec:4_role_actions}
Character actions constitute the majority of story progression. 
Instead of relying on a fixed action space, our system utilizes natural language to describe action details, enabling open-domain behaviors. 

Actions are either proactive or reactive. During their turn as an initiator, characters proactively plan and execute actions based on their goals, states, and information about others. When designated as action targets, characters should respond accordingly.

\begin{figure}
    \centering
    \includegraphics[width=0.6\linewidth]{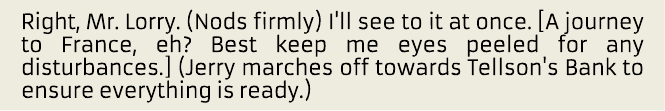}
    \caption{An example of role action $d_i$. Thoughts are enclosed in [ ]. Actions are enclosed in ( ). The character's dialogue is presented without any framing.}
    \label{fig:role_action_example}
\end{figure}

The actions that the initiator can take are classified into the following types based on the interaction targets:

\begin{itemize}
    \item{\textbf{Character Interaction}} Initiators can engage with characters either individually or in groups. Accessible characters include main characters and non-player characters (NPCs). NPCs, such as street vendors, are characters that are not assigned specific role agents. When an NPC is approached, the system creates a temporary, memoryless NPC agent to manage the interaction.
    \item{\textbf{Environmental Interaction}} Initiators can perform actions involving the environment to gather information or complete tasks, such as investigating the surrounding area.
    \item{\textbf{No Interaction}} If not specifying a target for interaction, initiators can engage in solitary activities, such as reading.
\end{itemize}

\paragraph{Memory}
To support long-term simulation, we design a long and short-term memory mechanism following generative agents~\citep{park2023generative}.

\begin{itemize}
    \item{\textbf{Short-Term Memory (STM)}} STM stores recent events and complete dialogue details $\langle...,d_{i-1},d_i\rangle$ up to a capacity limit, allowing for immediate responses in current scenarios.
    \item{\textbf{Long-Term Memory (LTM)}} LTM stores condensed summaries of memories exceeding the STM limit. When STM reaches capacity, older memories are abstracted to LTM (e.g. $m_i = \text{summarize}(d_i)$) and stored, available for retrieval when needed.
\end{itemize}

\subsubsection{World Agent}
The world agent serves as an indispensable component of this system, handling all tasks beyond character-to-character interactions, including environmental management and story outline processing.

\paragraph{Attributes}
The world agent primarily requires a basic worldview. It refers to a fundamental description of the virtual world, encompassing its primary characteristics and core settings. This overview provides agents with a basic framework for rapid comprehension of the world's essential features.

\paragraph{Actions}
The world agent maintains the virtual environment and primarily responds to actions taken by character agents. Its functionalities include:

\begin{itemize}
    \item{\textbf{Environmental Responses}} When character agents interact with the environment, the world agent generates outcomes based on the worldview settings and relevant information of the current location. For example, if a character attempts to break through a door, the action is more likely to succeed in a common village setting but may fail in a heavily guarded castle.
    \item{\textbf{Event Generation and Updates}} The world agent manages global events within the system. While stories are driven by conflicts, role agents operating without guidance may lose clear goals and exhibit repetitive behaviors. Therefore, generating stimulating events significantly enhances the story's appeal. When event stimulation is required by user settings, the World Agent generates conflict-rich events based on background settings and updates them according to the characters' real-time actions.
\end{itemize}

\subsubsection{Map}
\mas implements a discrete map to introduce spatial relationships in the environment. The map equips natural language descriptions for key locations and a weighted undirected graph for distances between locations. 
\paragraph{Location Profiles} Each location has a unique name, a brief description (appearance, atmosphere, history), and optional detailed information (local customs, special items).

\paragraph{Distance Network} Distances between locations are represented through a weighted undirected graph. Characters can move between adjacent locations or traverse longer paths by consuming specified time units. In \mas we specify one scene as a time unit.

\subsection{Simulation Implementation}
\label{sec:simulation_implementation}
\paragraph{Scene and Settlement} 
Prior to each scene's commencement, the world agent selects participating characters who must share a common location, ensuring narrative focus and interaction coherence. 

Each scene comprises multiple rounds, where characters have several action opportunities as the initiator. Rather than following a fixed sequence, the world agent dynamically determines the initiator based on character states, and narrative development requirements. During his/her round, the initiator makes the plan and action based on his/her objective and current situation. The available action types are explained in \S\ref{sec:4_role_actions}.

The system evaluates scene completion through the action records. Upon scene conclusion, characters may opt to move to another location. The world agent updates current events based on recent developments, and in script mode, provides guidance for the next narrative phase. Movement calculations are performed for traveling characters, with those completing their designated travel rounds arriving at their destinations.

\paragraph{Controllable Story Generating}
Based on practical requirements, users may wish to either observe characters' spontaneous actions or maintain direct control over the storyline. To accommodate these diverse needs, this system implements two operational modes with distinct feedback mechanisms: Free Mode and Script Mode.

\textbf{Script Mode} incorporates user-defined scripts to guide character actions, generating detailed behaviors while adhering to the script outline. At the start of the simulation, the system split given $script$ into critical acts $\langle act_1,act_2,\ldots\rangle$. Within the simulation, the system checks the progress and makes instructions to role agents based on current $act_i$, maintaining the narrative consistency with each session.

In \textbf{Free Mode}, characters have complete autonomy, acting based on their established settings and characteristics. Users can also set initial incidents, enhancing the drama and characters' engagement. Events update in real-time with simulation progress.

\subsection{Data Preparation}
\label{sec:data_preparation}
\paragraph{General Information Extraction}
We provide an automatic extraction method based on incremental updating inspired by~\citet{yuan2024evaluating}. 
The original text is first segmented into chunks. Target characters are allocated with an initial profile. Then, we traverse through chunks and recursively update the character information, including character profiles and relationships with others. This information is finally organized into structured data used for agent construction.

We conduct information extraction from six Chinese works and ten English works, producing a total of 453 presets. The presets mainly include the outline and the information about the present characters of a certain act in the book.

\paragraph{Enriching \mas with Worldview Data}
Novels, especially fictions, often contain implicit, unstructured knowledge that is not presented in a uniform format but can be inferred from the context~\citep{wang2023open}. For instance, in the world of \textit{Harry Potter}, ordinary people should be unaware of the existence of wizards. Any violations of established settings during the simulation would disrupt the narrative immersion of the experience. Therefore, a comprehensive database of the target work's worldview elements is essential.

\begin{figure}
    \centering
    \includegraphics[width=0.6\linewidth]{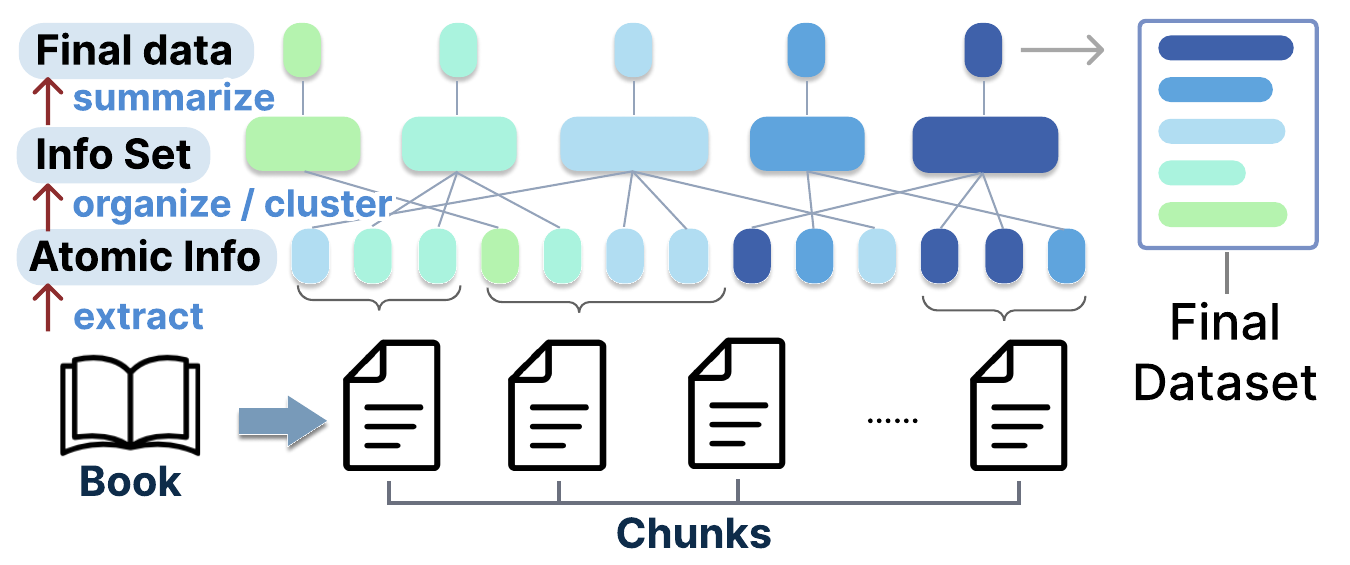}
    \caption{The procedure of building worldview data. We extract atomic facts from each chunk, then filter, cluster and summarize them to construct the final dataset.}
    \label{fig:booksettings_extraction_method}
\end{figure}

We propose a term-based extraction method to consolidate related settings. During simulation, these settings are incorporated into the prompts as references if certain terms are mentioned or the context is strongly related to specific settings, thereby enhancing the environmental immersion and maintaining consistent world-building throughout the simulation.

Each extracted setting contains four attributes:
\begin{itemize}
    \item{\textbf{Term}}: The object of description (e.g., ``Invisibility Cloak'', ``White Walkers''). Terms may be empty strings when describing general social atmospheres or universal facts within the fictional world.
    \item{\textbf{Nature}}: The category of the setting. This might include ``artifact'', ``social norm''.
    \item{\textbf{Detail}}: Comprehensive textual description of the setting. For a specific term, the detail describes its characteristics.
    \item{\textbf{Source}}: The chapter where the setting is extracted, enabling distinction between different timelines within the work.
\end{itemize}

We extract information in four steps. The procedure is illustrated in Figure~\ref{fig:booksettings_extraction_method}. 
First, we segment chapters into manageable chunks for processing. Next, we analyze each chunk to extract relevant worldview knowledge by equipping LLMs to identify elements in the chunk as well as their natures and details. Then, we filter these elements to remove character actions and common-sense information that could hinder the simulation. Finally, we cluster and consolidate similar records, eliminating redundancy while retaining crucial details.
Each element is tagged with its source chapter, facilitating both verification and timeline control in subsequent applications. 
We have collected 9912 settings from the same source materials mentioned above, 44.8\% in Chinese and 55.2\% in English. Details about extracted settings are presented in Appendix~\ref{sec:booksetting_detail}.

\begin{table*}[t]
\centering
\small
\begin{tabular}{ll|ccccc|ccccc}
\toprule
\multirow{2}{*}{\textbf{Model}} &
\multirow{2}{*}{\textbf{Method}} &
\multicolumn{5}{c}{\textit{with script}} & \multicolumn{5}{c}{\textit{without script}} \\
 &  
&  \textbf{An.} & \textbf{CF.} & \textbf{IS.} & \textbf{WQ.} & \textbf{SQ.} 
&  \textbf{An.} & \textbf{CF.} & \textbf{IS.} & \textbf{WQ.} & \textbf{Cr.} \\
\midrule
\rowcolor[rgb]{ .949,  .953,  .961} 
\multicolumn{12}{c}{ \textit{close-sourced models}}\\
\multirow{2}{*}{\texttt{gpt-4o-mini}} &
BW vs Direct.          
& 75.3 & 84.1 & 95.6 & 91.2 & 86.9 
& 91.3 & 73.9 & 98.5 & 91.3 & 87.0
\\
& \quad\;\,\,  vs HW     
& 69.5 & 73.9 & 82.6 & 65.2 & 60.8 
& 56.5 & 60.9 & 91.3 & 78.3 & 73.9
\\
\midrule

\multirow{2}{*}{\texttt{gemini-2}} &
BW vs Direct.         
& 93.5 & 82.6 & 87.0 & 95.7 & 84.8
& 95.6 & 95.7 & 100.0 & 98.5 & 89.8
\\
& \quad\;\,\,  vs HW    
& 89.1 & 65.2 & 89.1 & 97.8 & 60.9
& 94.2 & 87.0 & 97.1 & 100.0 & 97.1
\\
\midrule

\multirow{2}{*}{\texttt{qwen-plus}} &
BW vs Direct.         
& 87.0 & 73.9 & 91.3 & 87.0 & 73.9
& 95.7 & 82.6 & 87.0 & 95.7 & 91.3
\\
& \quad\;\,\,  vs HW    
& 82.6 & 56.5 & 89.1 & 82.6 & 45.7
& 82.6 & 69.6 & 100.0 & 91.3 & 95.7
\\
\midrule

\rowcolor[rgb]{ .949,  .953,  .961} 
\multicolumn{12}{c}{ \textit{open-source models}}\\

\multirow{2}{*}{\texttt{deepseek-v3}} &
BW vs Direct.         
& 80.8 & 88.5 & 97.1 & 88.5 & 73.1
& 91.3 & 91.3 & 100.0 & 95.7 & 95.7
\\
& \quad\;\,\,  vs HW    
& 76.9 & 80.8 & 92.3 & 92.3 & 53.8
& 82.6 & 87.0 & 95.7 & 95.7 & 91.3
\\
\midrule

\multirow{2}{*}{\texttt{Llama-3.3-70B}} &
BW vs Direct.         
& 46.2 & 50.7 & 52.2 & 69.2 & 42.3
& 34.6 & 42.3 & 53.8 & 34.6 & 47.8
\\
& \quad\;\,\,   vs HW    
& 26.9 & 42.3 & 53.8 & 39.1 & 46.2
& 46.2 & 49.2 & 50.7 & 69.2 & 42.3
\\
\midrule

\multirow{2}{*}{\texttt{qwen2.5-72B}} &
BW vs Direct.         
& 60.9 & 47.8 & 56.5 & 56.5 & 34.8
& 61.5 & 65.4 & 84.6 & 96.2 & 61.5
\\
& \quad\;\,\,  vs HW    
& 62.3& 52.2 & 69.6 & 73.9 & 39.1
& 65.4 & 61.5 & 88.5 & 96.2 & 69.2
\\

\bottomrule
\end{tabular}
\caption{The win rate (\%) of \mas (BW) against baseline methods. Direct. and HW denote direct generation and HoLLMwood respectively.}
\label{tab:result_winrate_4o}
\end{table*}

\section{Experiments}
In this section, we evaluate the effectiveness of \mas via story generation.
We focus on the quality of the generated stories and conduct extensive experiments.  

\subsection{Evaluation Metrics}

Since giving a concrete score for a story is quite a difficult task even for human evaluators, we adopt pairwise comparison across multiple dimensions between the stories generated by different methods following previous work~\citep{chen2024hollmwood}. Within this framework, outputs from any two methods undergo paired comparison, with LLMs determining superior performance. We demonstrate the credibility of this method by comparing it with human evaluation~\ref{app:human_eval}.

We conduct experiments and evaluations for two scenarios: story generation with a given outline and without an outline. The two scenarios require different evaluation dimensions. For the former, we assess \textbf{Storyline Quality (SQ.)}, which measures the system's adherence to the given outline. For the latter, we calculate \textbf{Creativity (Cr.)}, evaluating whether the system can produce innovative and refreshing new stories. Additionally, there are four general evaluation metrics that apply generally to both scenarios, making a total of five metrics assessed in each type of experiment:

\textbf{Anthropomorphism (An)}: The effectiveness of attributing human characteristics to non-human entities while preserving their original nature.

\textbf{Character Fidelity (CF)}: The consistency of characters' behaviors and actions with their established traits and backgrounds.

\textbf{Immersion and Setting (IS)}: The ability to create a convincing and engaging story world through environmental and atmospheric details.

\textbf{Writing Quality (WQ)}: The technical execution of writing mechanics and style that serves the narrative.

\subsection{Experiment Setup}

\paragraph{Baselines}
We compare stories generated via three methods: 
\textit{1)} Direct generation, which directly prompts LLMs with all the processed book data, 
\textit{2)} HoLLMwood~\citep{chen2024hollmwood}, utilizes LLMs to replicate the human story-writing process. The writer agent refines and breaks down the outline based on the editor agent's feedback, while the actor agents adopt character roles to flesh out the story.
and \textit{3)} \mas.

\paragraph{Models}
For simulation, 
we adopt both open-sourced models and close-sourced models as the base model for role agents and the world agent. 
During the experiments, we keep the number of dialogue turns consistent across methods. Each experiment simulates 2 to 4 scenes, averaging a total of 4,230 words. The final results are evaluated using \texttt{gpt-4o-2024-08-06}.

\subsection{Evaluation Results}
\label{sec:7_eval_baseline}
Table~\ref{tab:result_winrate_4o} illustrates the win rates of \mas against two baseline methods across five evaluation metrics. \mas consistently outperforms the direct generation baseline across all evaluated metrics and models, with particularly strong performance in Immersion. Against HoLLMwood, while \mas retains superiority in Immersion and Character Fidelity, it faces challenges in Storyline Quality and Writing Quality.

On most models, \mas demonstrates significant advantages over baseline approaches. However, when using \texttt{Llama-3.3-70B}, \mas underperforms direct generation and HW in certain dimensions. 
This performance gap can be attributed to two factors. First, multi-agent simulation is inherently more complex than direct generation, requiring advanced capabilities in handling structured outputs and following complicated instructions. 
Second, our experiments partially involve Chinese data, which is not an advantage of \texttt{Llama-3.3-70B}.

\begin{table}[h]
\centering
\small
\begin{tabular}{lccccc}
\toprule
\textbf{Function} &  \textbf{An.} & \textbf{CF.} & \textbf{IS.} & \textbf{WQ.} & \textbf{SQ.}  \\
\midrule
\rowcolor[rgb]{ .949,  .953,  .961} 
\multicolumn{6}{c}{ \textit{with script}}\\
w \textit{vs} w/o Scene
& 84.1 & 63.7 & 88.4 & 86.9 & 76.8 \\
w \textit{vs} w/o Env.
& 55.3 & 60.5 & 81.6  & 71.1 & 52.6 \\
w \textit{vs} w/o Set.
& 50.7 & 47.8 & 76.8  & 58.0 & 54.7 \\

\midrule
\textbf{Function} &  \textbf{An.} & \textbf{CF.} & \textbf{IS.} & \textbf{WQ.} & \textbf{Cr.}   \\
\midrule
\rowcolor[rgb]{ .949,  .953,  .961} 
\multicolumn{6}{c}{ \textit{without script}}\\
w \textit{vs} w/o Scene
& 87.0 & 60.9 & 97.1 & 92.7 & 73.9 \\
w \textit{vs} w/o Env.
& 48.5 & 56.5 & 94.2  & 82.6 & 52.2 \\
w \textit{vs} w/o Set.
& 52.2 & 43.5 & 73.9  & 44.8 & 56.5 \\

\bottomrule
\end{tabular}
\caption{The win rate (\%) from the ablation study, Comparing \mas with full functionality against versions without specific functions, where Env. refers to environment response and Set. refers to the settings extracted from the book.}
\label{tab:result_winrate_ablation}
\end{table}

\begin{table*}[h]
\centering
\scriptsize
\renewcommand{\arraystretch}{1.1} 
\setlength{\tabcolsep}{8pt}
\begin{tabular}{p{1.5cm}p{6.5cm}p{6.5cm}}
    \toprule

    \multicolumn{3}{l}{\footnotesize\textbf{Enviroment and NPC Responses}} \\
    \midrule
    \textbf{Type} & \textbf{Enviroment} & \textbf{NPC} \\

    \textbf{World Agent} & \hl{(Characters broke the seal)} As they stood transfixed, \greenemph{the mist before them slowly parted like a ethereal curtain, unveiling a sight} that stole their breath away. There, nestled in the heart of the...
    & \hl{(As Tellson's Bank Manager)} With a steadfast nod, \greenemph{the bank manager acknowledged Mr. Lorry's request}. Understanding the gravity of the situation, he made a swift motion to gather the nece-..
    \\
    \midrule
    
    \multicolumn{3}{l}{\footnotesize\textbf{Memories and States help maintain Context Consistency}} \\
    \midrule
    \textbf{Reference} & \textbf{Memory} & \textbf{State} \\

    \textbf{Role Agent} & \hl{(Arya possesses memories of the Red Wedding.)} [Jaime's words sent a shockwave through her heart, \greenemph{yet hatred continued to burn like wildfire within her.}]
    Arya clenched her jaw tightly and said...
    & \hl{(State: Unconscious)} \greenemph{Bran Stark's consciousness floated in the darkness}, where distant voices echoed beyond his reach, leaving his attempts to respond trapped in the void. \\
    \midrule

    \multicolumn{3}{l}{\footnotesize\textbf{Settings provide necessary background knowledge}} \\
    \midrule
    \textbf{Reference} & \textbf{without settings} & \textbf{with settings} \\

    \textbf{\mas} & Somewhere on Solaris, the ocean rippled gently under the \redemph{moonlight}, like a veil of \redemph{silver} gossamer. The air was thick with unknown whispers, as if dreams from the abyss had found their voice.
    & \hl{(Solaris is a planet orbiting two suns, one red and one blue.)} The ocean undulated \greenemph{like a living entity} under the strange radiance of \greenemph{the twin suns}. In the interplay of \greenemph{red and blue celestial light}...  \\
    \midrule
    
    \multicolumn{3}{l}{\footnotesize\textbf{Convert Simulation Records to Novel-style Stories}} \\
    \midrule
    \textbf{Stage} & \textbf{Simulation Records of \mas} & \textbf{Novel-Style Story} \\

    \textbf{\mas} & 
    Jarvis Lorry: "Jerry, I need you to ensure that all necessary...
    
    Jerry Cruncher: "Right, Mr. Lorry. (Nods firmly) I'll see to it.
    
    Jarvis Lorry: "With a steadfast nod, the bank manager...
    & In the bustling heart of London, within the well-worn walls of Tellson’s Bank, Mr. Lorry sat in quiet contemplation, his mind occupied with thoughts...
\\
        
    \bottomrule
    \end{tabular}
    \caption{Some examples of intermediate outputs in \mas. The \redemph{red text} is used to indicate situations where the \mas output does not meet expectations, while the \greenemph{green text} represents the correct output.}
    \label{tab:case_study}
\end{table*}

\subsection{Ablation Study}
We conducted an ablation study and comparative analysis on the primary features of \mas. Using \texttt{gpt-4o-mini-2024-07-18} as the base model for the agents, adopting same presets with \S~\ref{sec:7_eval_baseline}, we compared the output quality of \mas with and without specific features enabled, calculating the win rate for each dimension. The final results are presented in the Table~\ref{tab:result_winrate_ablation}.

The results are basically in line with expectations. Removing environmental output significantly affect the sense of immersion, which in turn reduced writing quality, but it did not have much impact on the quality of the storyline. On the other hand, disabling the Scene mode results in a decline in quality across all dimensions, with the most significant impact on the storyline.

\subsection{Discussion}
We display some interesting examples of \mas outputs in Table~\ref{tab:case_study}. 
In most cases, \mas can process information correctly, yielding appropriate results when the agents interact with the environment and initiate interactions. The system has the capability to maintain and utilize long-term memories and states. We also demonstrated the impact of introducing worldview settings. Prior to incorporating these settings, the model generated environment responses based on real-world common sense, leading to descriptions such as \dq{moonlight} and \dq{silver}, which are inconsistent with the worldview of \textit{Solaris}. 

\section{Conclusion}
In this paper, we presented \mas, a comprehensive system that transforms static literary works into dynamic, interactive environments. Our approach differs from previous work by focusing specifically on reproducing the unique worldviews, geographical settings, and interpersonal dynamics that make these literary works compelling.
Experiment results show that \mas successfully creates high-quality narratives by building immersive book-based societies. 
Our approach demonstrates high scalability and broad applicability across various scenarios. We hope this research will further advance the development of multi-agent technologies and character simulation techniques.

\section*{Limitations}
To prioritize generalizability and openness, \mas adopts a highly simplified representation of interactive environments. This trade-off leads to reduced performance compared to systems specifically designed for particular works or scenarios. For example, it is nearly impossible to complete a full game of Werewolf (a social deduction game) in \mas.

Additionally, current research in role-playing technology predominantly focuses on one-on-one chat between users and characters, with limited attention to characters' decision-making processes in realistic environments. This research gap results in characters exhibiting indecisive behaviors when faced with complex situations. Addressing this limitation requires further advancement in role-playing technology, particularly in enhancing characters' capabilities in complex, multi-agent scenarios.

\section*{Ethics Statement}
In this paper, we introduce \mas . The development and use of \mas are guided by ethical principles to ensure responsible and beneficial outcomes. 
We extracted sample data using six Chinese novels and ten English novels, primarily conducting experiments with data extracted from the \textit{A Song of Ice and Fire} series. We affirm that our research is conducted for academic and non-commercial purposes only. The use of these texts is solely for the development and evaluation of our models in natural language processing tasks, aimed at advancing scientific knowledge in the field.

\paragraph{Use of Human Annotations} In our research, we conducted a comparative analysis of human and LLM evaluation results to validate the reliability of LLM evaluators. Our human annotators were university students with deep familiarity with the source novels. To ensure ethical research practices, we provided compensation well above local minimum wage standards and maintained full transparency regarding the purpose and application of their annotations. We obtained informed consent from all participants for the use of their contributions in our research. Throughout the process, we prioritized the protection of annotators' privacy rights, maintaining strict confidentiality protocols to create an ethical and respectful research environment.

\paragraph{Risk} Our method is used to build a book-based interactive society. Firstly, it is constrained by limitations; it may not fully capture the complexities of human interactions and narrative depth inherent in traditional storytelling. Secondly, this method could potentially be misused for unintended purposes, such as generating misleading or harmful content, which raises ethical concerns around its application and necessitates careful oversight and regulation.

We encourage the responsible use of \mas for educational, entertainment, and creative purposes while discouraging any harmful or malicious activities.

\bibliography{custom,anthology}
\bibliographystyle{acl_natbib}

\clearpage

\appendix
\section{Consistency with Human Evaluation}
\label{app:human_eval}
To validate the reliability of our model-based evaluation approach, we conduct a comprehensive agreement analysis between model assessments and human assessments. We recruit 5 human annotators (Fans of the corresponding work) to evaluate the outputs from our proposed method and the baseline approach across five dimensions.

For each comparison, both human annotators and our model are asked to indicate which method performs better on each dimension. To ensure evaluation quality, we randomly sample Y pairs of outputs from both methods for assessment. To quantify the agreement between model and human judgments, we employ Cohen's Kappa coefficient ($\kappa$), a metric for measuring inter-rater reliability while accounting for chance agreement.
\begin{table}[h]
\centering
\smallskip
\begin{tabular}{cccccc}
\toprule
 &  An. & CF. & IS. & WQ. & SQ.  \\
\hline
$\kappa$       
& 0.786 & 0.688 & 0.637 & 0.781 & 0.731 \\

\bottomrule
\end{tabular}
\caption{The Cohen's Kappa between human evaluation and model evaluation.}
\label{tab:a_cohen_kappa}
\end{table}

The Cohen's Kappa coefficients between our model's judgments and human evaluations are presented in Table~\ref{tab:a_cohen_kappa}. The results indicate that there is a high level of consistency between human evaluation and machine evaluation, and the results of machine evaluation are sufficient to reflect the actual effectiveness of the method.

\section{Dataset Details}
\label{sec:booksetting_detail}
\begin{table*}[h]
\centering
\setlength{\tabcolsep}{8pt}
\begin{tabular}{p{6cm}cccc}
    \toprule
     \textbf{Title} 
     & \textbf{Language} 
     & \textbf{\#Settings} 
     &\textbf{\#Chapters} 
     &\textbf{ \#Words}\\
    \midrule
Dracula 
& en & 1113
& 107 & 165453 \\

Othello 
& en & 275
& 6 & 98558 \\

The Adventures of Tom Sawyer 
& en & 753
& 47 & 81971 \\

Paradise Lost 
& en & 936
& 27 & 270030 \\

A Study in Scarlet (Sherlock Holmes, \#1) 
& en & 155
& 20 &  67440 \\

Alice’s Adventures in Wonderland - Through the Looking-Glass 
& en & 169
& 46 & 199252 \\

Around the World in Eighty Days 
& en & 334
& 49 & 110435 \\

Don Quixote 
& en & 121
& 138 &  823174 \\

Treasure Island 
& en & 299
& 63 & 82327 \\

Uncle Tom’s Cabin 
& en & 538
& 57 & 387799 \\

A Song of Ice and Fire (Part)
& zh & 252
& 56 & 300628\\

Three-Body
& zh & 1684
& 62 &876196 \\

Ball Lightning 
& zh & 322
& 36 & 180658 \\

The True Story of Ah Q
& zh & 56
&11 & 70856 \\

The Deer and the Cauldron
& zh & 1816
& 55 & 1243532 \\

Solaris 
& zh & 319
& 18 &144972 \\

    \bottomrule
\end{tabular}
\caption{The detailed information about extracted worldview settings.}
\label{tab:Booksetting_detail}
\end{table*}

\begin{table*}[h]
\centering
\renewcommand{\arraystretch}{1.2} 
\setlength{\tabcolsep}{8pt}
\begin{tabular}{p{2.5cm}p{2cm}p{8cm}p{2cm}}
    \toprule
     \textbf{Term} & \textbf{Nature} & \textbf{Detail} &\textbf{ Source}\\
    \midrule
    Abyss & location& The abyss serves as a gulf between heaven and hell, embodying the separation of these two realms. & 7\_Book\_2
    \\
    \hline
     Altar & ritual site  &  The practice of placing altars for deities beside altars for the God of monotheistic faiths, indicating a clash of religious beliefs.", "A raised structure used for offerings and sacrifices, often associated with worship and remembrance of divine encounters. & 5\_Book\_1, 21\_Book\_11   \\
    \hline
      Bellerophon & character & Bellerophon is a figure representing the consequences of hubris, as he attempted to ascend to heaven but faced dire repercussions for his ambition. & 15\_Book\_7 \\
    \hline
      Bridge of Hell &construction  & A monumental bridge is constructed over the Abyss, symbolizing the connection between Hell and the earthly realm, allowing for the passage of entities between these worlds. & 19\_Book\_10 \\
    \hline
       Empyrean & realm & The highest and most divine part of the universe, often associated with the presence of the divine and ultimate reality. & 9\_Book\_3 \\
    \hline
       First Fruits &  ritual& The presentation of the first harvest to the divine is a sacred act, symbolizing gratitude and the acknowledgment of a higher power's role in agriculture. & 22\_Book\_11 \\
    \bottomrule
\end{tabular}
\caption{Some examples of settings, extracted from \textit{Paradise Lost}.}
\label{tab:Booksetting_sample}
\end{table*}

We adopt the extraction method on a diverse corpus of 16 novels, comprising 10 English and 6 Chinese works. The experiment yields a total of 9,142 setting entries, with 4,449 entries extracted from Chinese novels and 4,693 from English novels. The sources and relevant information are listed in Table~\ref{tab:Booksetting_detail}
We display some example settings from our dataset in Table~\ref{tab:Booksetting_sample}.

\section{Retrieval Augmented Role-Playing Agent}
\label{sec:appendix_rpa}
The construction of role agents aims to optimize the output $a$ of a query $q$ directed to the agent $A(R, M)$, where $R$ refers to character information and $M$ means base model. In this research, we follow the method introduced in ChatHaruhi~\citep{li2023chatharuhi}. This involves leveraging three key sources: 

\paragraph{Character Profile ($profile$)} A concise description of the character, including their appearance, personality traits, and background. Due to its brevity, the entire profile can be retained and included in the model's input without modification.

\paragraph{Original Text Excerpts ($text$)} A set of character-related excerpts from the source material, typically represented as a list $\{t_1, t_2, \ldots, t_n\}$ (e.g., all the dialogues of a certain character in the original work). The total volume of text is often too large to be fully input into the model. 

\paragraph{Historical Records ($memory$)} 
As the simulation progresses, character agents continuously generate new action records~$\{m_1, m_2, \ldots, m_k\}$. 

For $memory$ and $text$, a vector similarity-based retrieval method is employed to identify the top $k$ relevant records. The retrieval mechanisms are similar. Consider the retrieval of $text$, each excerpt $t_i$ is encoded into a fixed-length vector $v_i$ using a pre-trained text embedding model, and the vectors are stored in a vector database. When a query $q$ is issued, it is similarly transformed into a vector representation $v_q$ using the same embedding model. The similarity between $q$ and each text excerpt $t_i$ is then computed using the cosine similarity metric:

\[
\mathrm{similarity}(q, t_i) = \cos(v_q, v_i) = \frac{v_q \cdot v_i}{|v_q||v_i|}.
\]

Based on the computed similarity scores, the top $k$ most similar excerpts are selected as references for role-playing:

\[
\{t_{i_1}, t_{i_2}, \ldots, t_{i_k}\} = \underset{i \in \{1, 2, \ldots, N\}}{\mathrm{argmax}} \, \mathrm{similarity}(q, v_i).
\]

The final prompt for the role-playing agent consists of four components: the system’s role-playing instructions, the $profile$, the retrieved memory $\{m_{i_1}, m_{i_2}, \ldots, m_{i_k}\}$, and the retrieved text excerpts $\{t_{i_1}, t_{i_2}, \ldots, t_{i_k}\}$. This prompt is then fed into the language model $M$, which generates the response $a$ for the given query $q$.

\section{Prompts}
\begin{table*}[h]
    \centering
    \scriptsize
    \renewcommand{\arraystretch}{1.1} 
    \setlength{\tabcolsep}{8pt}
    \begin{tabular}{p{16cm}}
        \toprule
        \textbf{Role Agent Prompt for Planning} \\
        \hline
        
You are \{role\_name\}. Your nickname is \{nickname\}. Based on your goal and other provided information, you need to take the next action.
\\ \hspace*{\fill} \\
\#\# Action History 

\{history\}
\\ \hspace*{\fill} \\

\#\# Your profile

\{profile\}
\\ \hspace*{\fill} \\

\{world\_description\}
\\ \hspace*{\fill} \\

\#\# Your goal

\{goal\}
\\ \hspace*{\fill} \\

\#\# Your status

\{status\}
\\ \hspace*{\fill} \\

\#\# Other characters with you; currently, you can only interact with them

\{other\_roles\_info\}
\\ \hspace*{\fill} \\

\#\# Roleplaying Requirements

1. **Output Format:** Your output, "detail," can include **thoughts**, **speech**, or **actions**, each occurring 0 to 1 time. Use [] to indicate thoughts, which are invisible to others. Use () to indicate actions, such as “(silence)” or “(smile),” which are visible to others. Speech needs no indication and is visible to others.

   - Note that **actions** must use your third-person form, {nickname}, as the subject.  

   - For speech, refer to the speaking habits outlined in: {references}.  
\\ \hspace*{\fill} \\

2. **Roleplay \{nickname\}:** Imitate his/her language, personality, emotions, thought processes, and behavior. Plan your responses based on their identity, background, and knowledge. Exhibit appropriate emotions and incorporate subtext and emotional depth. Strive to act like a realistic, emotionally rich person.  

   The dialogue should be engaging, advance the plot, and reveal the character's emotions, intentions, or conflicts.  

   Maintain a natural flow in conversations; for instance, if the prior dialogue involves another character, **avoid repeating that character's name**.  

   - You may reference the relevant world-building context: {knowledges}.  
\\ \hspace*{\fill} \\

3. **Concise Output:** Each paragraph of thoughts, speech, or actions should typically not exceed 40 words.  
\\ \hspace*{\fill} \\

4. **Substance:** Ensure your responses are meaningful, create tension, resolve issues, or introduce dramatic twists.  
\\ \hspace*{\fill} \\

5. **Avoid Repetition:** Avoid repeating information from the dialogue history, and refrain from vague or generic responses. Do not “prepare,” “ask for opinions,” or “confirm”; instead, act immediately and draw conclusions.
\\ \hspace*{\fill} \\

Return the response following JSON format.
It should be parsable using eval(). **Don't include ```json**. Avoid using single quotes '' for keys and values, use double quotes.
\\ \hspace*{\fill} \\

Output Fields:

'action': Represents the action, expressed as a single verb.

'interact\_type': 'role', 'environment', 'npc', or 'no'. Indicates the interaction target of your action. 

  - 'role': Specifies interaction with one or more characters. 
  
    - If 'single', you are interacting with a single character (e.g., action: dialogue).
    
    - If 'multi', you are interacting with multiple characters.
    
  - 'environment': Indicates interaction with the environment (e.g., action: investigate, destroy).
  
  - 'npc': Refers to interaction with a non-character in the list (e.g., action: shop).
  
  - 'no': Indicates no interaction is required.
  
'target\_role\_codes': list of str. If 'interact\_type' is 'single' or 'multi', it represents the list of target character codes, e.g., ["John-zh", "Sam-zh"]. For 'single', this list should have exactly one element.

'target\_npc\_name': str. If 'interact\_type' is 'npc', this represents the target NPC name, e.g., "shopkeeper."

'visible\_role\_codes': list of str. You can limit the visibility of your action details to specific group members. This list should include 'target\_role\_codes'.

'detail': str. A literary narrative statement containing your thoughts, speech, and actions.
\\

        \bottomrule
    \end{tabular}
    \caption{Planning prompt for role agent.}
\end{table*}

\begin{table*}[h]
    \centering
    \small
    \renewcommand{\arraystretch}{1.1} 
    \setlength{\tabcolsep}{8pt}
    \begin{tabular}{p{16cm}}
        \toprule
        \textbf{World Agent Prompt for Environmental Interaction} \\
        \hline
        
You are an Enviroment model, responsible for generating environmental information. Character \{role\_name\} is attempting to take action \{action\} at \{location\}.
\\ \hspace*{\fill} \\

Based on the following information, generate a literary description that details the process and outcome of the action, including environmental details and emotional nuances, as if from a narrative novel. Avoid using any system prompts or mechanical language. Return a string.
\\ \hspace*{\fill} \\

\#\# Action Details

\{action\_detail\}
\\ \hspace*{\fill} \\

\#\# Location Details

\{location\_description\}
\\ \hspace*{\fill} \\

\#\# Worldview Details

\{world\_description\}
\\ \hspace*{\fill} \\

\#\# Additional Information

\{references\}
\\ \hspace*{\fill} \\

\#\# Response Requirements

1. The action may fail, but avoid making the action ineffective. Try to provide new clues or environmental descriptions.
\\ \hspace*{\fill} \\

2. Use a third-person perspective.
\\ \hspace*{\fill} \\

3. Keep the output concise, within 100 words. You serve as the Enviroment model, responding to the character's current action, not performing any actions for the character.
\\ \hspace*{\fill} \\

4. The output should not include the original text from the action detail but should seamlessly follow the action details, maintaining the flow of the plot.

\\

        \bottomrule
    \end{tabular}
    \caption{Prompt of making environment response for the world agent.}
\end{table*}

We provide the details of the prompt templates of \mas in this section.

\end{document}